\documentclass[10pt,twocolumn,letterpaper]{article}
\usepackage{multirow}
\usepackage{cvpr}
\usepackage{times}
\usepackage{epsfig}
\usepackage{graphicx}
\usepackage{amsmath}
\usepackage{amssymb}
\usepackage{multirow}
\usepackage{relsize,bbm}
\usepackage{enumitem, standalone}
\graphicspath{ {./images/} }
\usepackage{subcaption,bm,times,csquotes}
\usepackage{verbatim,url,soul}
\usepackage{color,floatrow}
\usepackage{gensymb}
\usepackage{pdfpages}
\usepackage[pagebackref=true,breaklinks=true,letterpaper=true,colorlinks,bookmarks=false]{hyperref}

\usepackage{fancyhdr}

\cvprfinalcopy 

\def\cvprPaperID{4442} 

\begin{document}

\title{Joint Face Detection and Facial Motion Retargeting for Multiple Faces}

\author{Bindita Chaudhuri$^1$$^*$, Noranart Vesdapunt$^2$, Baoyuan Wang$^2$\\
$^1$University of Washington, $^2$ Microsoft Corporation\\
$^1${\tt\small bindita@cs.washington.edu}, $^2${\tt\small \{noves,baoyuanw\}@microsoft.com}
}

\maketitle

\begin{abstract}
\vspace{-7.5pt}
{\let\thefootnote\relax\footnote{{$^*$Work primarily done during an internship at Microsoft.}}}Facial motion retargeting is an important problem in both computer graphics and vision, which involves capturing the performance of a human face and transferring it to another 3D character. Learning 3D morphable model (3DMM) parameters from 2D face images using convolutional neural networks is common in 2D face alignment, 3D face reconstruction etc. However, existing methods either require an additional face detection step before retargeting or use a cascade of separate networks to perform detection followed by retargeting in a sequence. In this paper, we present a single end-to-end network to jointly predict the bounding box locations and 3DMM parameters for multiple faces. First, we design a novel multitask learning framework that learns a disentangled representation of 3DMM parameters for a single face. Then, we leverage the trained single face model to generate ground truth 3DMM parameters for multiple faces to train another network that performs joint face detection and motion retargeting for images with multiple faces. Experimental results show that our joint detection and retargeting network has high face detection accuracy and is robust to extreme expressions and poses while being faster than state-of-the-art methods.
\end{abstract}
\begin{figure*}[t]
\centering
\includegraphics[width=1.0\linewidth]{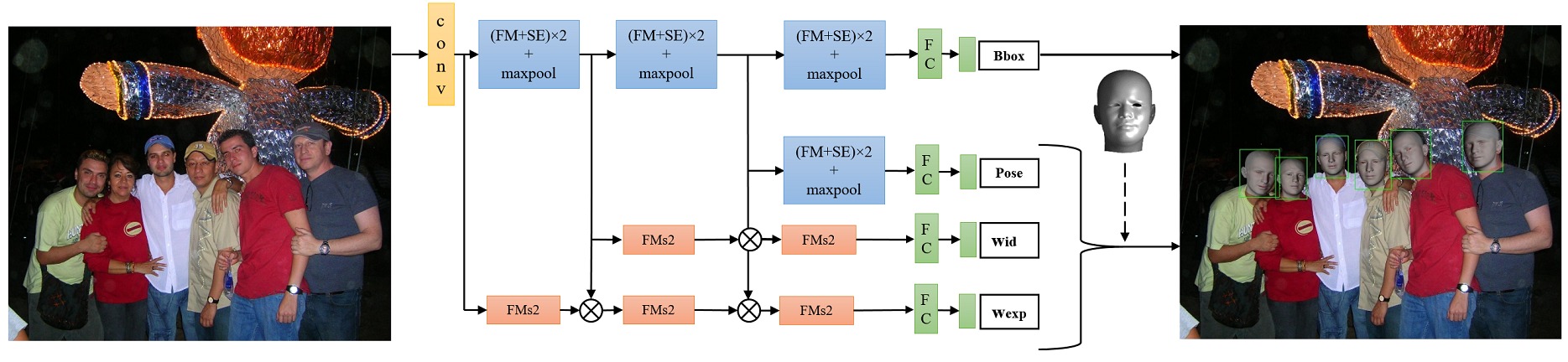}
\caption{Our end-to-end joint face detection and retargeting network is specifically tailored to incorporate scale prior and disentangling. The building blocks are Fire Module (FM) \cite{SqueezeNet} and squeeze-and-excitation (SE) \cite{SEnet} which are designed for real-time application. The multi-scale branch uses multiple slim FM with stride 2 on the last FM (FMs2) to allow concatenation. The multiplication of $Pose$, $w_{id}$, $w_{exp}$ with 3DMM generates 3D human mesh for every bounding box.}
\label{fig:teaser}
\end{figure*}

\vspace{-15pt}
\section{Introduction}
Facial gestures are an effective medium of non-verbal communication, and communication becomes more appealing through 3D animated characters. This has led to extensive research \cite{DDEregression,exprgen,hsieh2015unconstrained} in developing techniques to retarget human facial motion to 3D animated characters. The standard approach is to model human face by a 3D morphable model (3DMM)\cite{3DMM} and learn the weights of a linear combination of blendshapes that fits to the input face image. The learned ``expression" weights and ``head pose" angles are then directly mapped to semantically equivalent blendshapes of the target 3D character rig to drive the desired facial animation. Previous methods, such as \cite{DDEregression}, formulate 3DMM fitting as an optimization problem of regressing the 3DMM parameters from the input image. However, these methods require significant pre-processing or post-processing operations to get the final output.

Using deep convolution neural networks, recent works have shown remarkable accuracy in regressing 3DMM parameters from a 2D image. However, while 3DMM fitting with deep learning is frequently used in related domains like 2D face alignment\cite{3ddfa,adrian3Dfacealignment}, 3D face reconstruction\cite{CNN3DMMsynth,prn,dff,mobileface} etc., it hasn't been proven yet as an effective approach for facial motion retargeting. This is because 1) face alignment methods focus more on accurate facial landmark localization while face reconstruction methods focus more on accurate 3D shape and texture reconstruction to capture the fine geometric details. In contrast, facial retargeting to an arbitrary 3D character only requires accurate transfer of facial expression and head pose. However, due to the ambiguous nature of this ill-posed problem of extracting 3D face information from 2D image, both facial expression and head pose learned by those methods are generally \textit{sub-optimal} as they are not well decoupled from other information like identity. 2) Unlike alignment and reconstruction, retargeting often requires real-time tracking and transfer of the facial motion. However, existing methods for alignment and reconstruction are highly memory intensive and often involve complex rendering of the 3DMM as intermediate steps, thereby making these methods difficult to deploy on light-weight hardware like mobile phones. 

It is important to note that all previous deep learning based 3DMM fitting methods work on a single face image assuming face is already detected and cropped. To support multiple faces in a single image, a straightforward approach is to run a face detector on the image first to detect the all face regions and then perform the retargeting operations on each face individually. Such an approach, however, requires additional execution time for face detection and the computational complexity increases linearly with the number of faces in the input image. Additionally, tracking multiple faces with this approach becomes difficult when people move in and out from the frame or occlude each other. In the literature of joint face detection and alignment, existing methods \cite{JointDA_cascade, JointDA_cascade_mtl, Jointalignment} either use a random forest to predict the face bounding boxes and landmarks or adopt an iterative two-step approach to generate region proposals and predict the landmark locations in the proposed regions. However, these methods are primarily optimized for regressing accurate landmark locations rather than 3DMM parameters.

To this end, we divide our work into two parts. In the first part, we propose a multitask learning network to directly regress the 3DMM parameters from a well-cropped 2D image with a single face; we call this as Single Face Network (SFN). Our 3DMM parameters are grouped into: a) identity parameters that contain the face shape information, b) expression parameters that captures the facial expression, c) pose parameters that include the 3D rotation and 3D translation of the head and d) scale parameters that links the 3D face with the 2D image. We have observed that pose and scale parameters require global information while identity and expression parameters require different level of information, so we propose to emphasize on high level image features for pose and scale and the multi-scale features for identity and expression. Our network architecture is designed such that different layers embed image features at different resolutions, and these multi-scale features help in disentangling the parameter groups from each other. In the second part, we propose a single end-to-end trainable network to jointly detect the face bounding boxes and regress the 3DMM parameters for multiple faces in a single image. Inspired by YOLO\cite{yolo} and its variants\cite{yolov2,yolov3}, we design our Multiple Face Network (MFN) architecture that takes a 2D image as input and predicts the centroid position and dimensions of the bounding box as well as the 3DMM parameters for each face in the image. Unfortunately, existing publicly available multi-face image datasets provide ground truth for face bounding boxes only and not 3DMM parameters. Hence, we leverage our SFN to generate the weakly labelled ``ground truth" for 3DMM parameters for each face to train our MFN. Experimental results show that our MFN not only performs well for multi-face retargeting but also improves the accuracy of face detection. Our main contributions can be summarized as follows:
\begin{enumerate}
\itemsep0em 
\item We design a multitask learning network, specifically tailored for facial motion retargeting by casting the scale prior into a novel network topology to disentangle the representation learning. Such network has been proven to be crucial for both single face and multiple face 3DMM parameters estimation.
\item We present a novel top-down approach using an end-to-end trainable network to jointly learn the face bounding box locations and the 3DMM parameters from an image having multiple faces with different poses and expressions.
\item Our system is easy to deploy into practical applications without requiring separate face detection for pose and expression retargeting. Our joint network can be run in real-time on mobile devices without engineering level optimization, e.g. only 39ms on Google Pixel 2.
\end{enumerate}

\vspace{-10pt}
\section{Related Work}

\subsection{2D Face Alignment and 3D Face Reconstruction}
Early methods like \cite{alignmentbytrees} used a cascade of decision trees or other regressors to directly regress the facial landmark locations from a face image. Recently, the approach of regressing 3DMM parameters using CNNs and fitting 3DMM to the 2D image has become popular. While Jourabloo et al. \cite{facealign3D2} use a cascade of CNNs to alternately regress the shape (identity and expression) and pose parameters, Zhu et al. \cite{3ddfa,pncc_new} perform multiple iterations of a single CNN to regress the shape and pose parameters together. These methods use large networks and require 3DMM in the network during testing, thereby requiring large memory and execution time. Regressing 3DMM parameters using CNNs is also popular in face reconstruction \cite{CNN3DMM,CNN3DMMsynth,cnnreconst,Tewari}. Richardson et al. \cite{Richardson} uses a coarse-to-fine approach to capture fine details in addition to face geometry. However, reconstruction methods also regress texture and focus more on capturing fine geometric details. For joint face alignment and reconstruction, \cite{prnet} regresses a position regression map from the image and \cite{nonlinear3dmm} regresses the parameters of a nonlinear 3DMM using an unsupervised encoder-decoder network. For joint face detection and alignment, recent methods either use a mixture of trees \cite{afwdatapaper} or a cascade of CNNs \cite{JointDA_cascade, JointDA_cascade_mtl}. In \cite{Jointalignment}, separate networks are trained to perform different tasks like proposing regions, classifying and regressing the bounding boxes from the regions, predicting the landmark locations in those regions etc. In \cite{hyperface}, region proposals are first generated with selective search algorithm and bounding box and landmark locations are regressed for each proposal using a multitask learning network. In contrast, we use a single end-to-end network to do join face detection and 3DMM fitting for face retargeting purposes.

\subsection{Performance-Based Animation}
Traditional performance capture systems (using either depth cameras or 3D scanners for direct mesh registration with depth data) \cite{AAM&depth,onlinemodeling,avataranimation_blendshapes} require complex hardware setup that is not readily available. Among the methods which use 2D images as input, the blendshape interpolation technique \cite{DDEregression,avataranimation_landmarks} is most popular. However, these methods require dense correspondence of facial points \cite{mocap} or user-specific adaptations \cite{realtime:animation:onthefly,Cao3d} to estimate the blendshape weights. Recent CNN based approaches either require depth input \cite{realtime:cnn:animation,SelfsupervisedCF} or regress character-specific parameters with several constraints \cite{exprgen}. Commercial software products like Faceshift \cite{faceshift}, Faceware \cite{faceware} etc. perform realtime retargeting but with poor expression accuracy \cite{exprgen}.

\subsection{Object Detection and Keypoint Localization}
In the literature of multiple object detection and classification, Fast RCNN \cite{rcnn} and YOLO \cite{yolo} are the two most popular methods with state-of-the-art performance. While \cite{rcnn} uses a region proposal network to get candidate regions before classification, \cite{yolo} performs joint object location regression and classification. Keypoint localization for multiple objects is popularly used for human pose estimation \cite{multiposenet,multipose} or object pose estimation \cite{object6d}. In case of faces, landmark localization for multiple faces can be done in two approaches: \textit{top-down approach} where landmark locations are detected after detecting face regions and \textit{bottom-up approach} where the facial landmarks are initially predicted individually and then grouped together into face regions. In our method, we adopt the top-down approach.

\section{Methodology}

\subsection{3D Morphable Model}

The 3D mesh of a human face can be represented by a multilinear 3D Morphable Model (3DMM) as
\begin{equation}
\mathcal{M} = \mathcal{V} \times \text{b}_{\text{id}} \times \text{b}_{\text{exp}}
\end{equation}
where $\mathcal{V}$ is the mean neutral face, $\text{b}_{\text{id}}$ are the identity bases and $\text{b}_{\text{exp}}$ are the expression bases. We use the face tensor provided by FacewareHouse \cite{facewarehouse} as 3DMM, where $\mathcal{V} \in \mathbb{R}^{11510 \times 3}$ denotes $11,510$ 3D co-ordinates of the mesh vertices, $\text{b}_{\text{id}}$ denotes 50 shape bases obtained by taking PCA over 150 identities and $\text{b}_{\text{exp}}$ denotes 47 bases corresponding to 47 blendshapes (1 neutral and 46 micro-expressions). To reduce the computational complexity, we manually mark 68 vertices in $\mathcal{V}$ as the facial landmark points based on \cite{afwdatapaper} and create a reduced face tensor $\hat{\mathcal{M}} \in \mathbb{R}^{204 \times 50 \times 47}$ for use in our networks. Given a set of identity parameters $w_\text{id} \in \mathbb{R}^{50 \times 1}$, expression parameters $w_\text{exp} \in \mathbb{R}^{47 \times 1}$, 3D rotation matrix $\mathbf{R} \in \mathbb{R}^{3 \times 3}$, 3D translation parameters $\mathbf{t} \in \mathbb{R}^{3 \times 1}$ and a scale parameter (focal length) $f$, we use weak perspective projection to get the 2D landmarks $\mathbf{P_{lm}} \in \mathbb{R}^{68\times2}$ as:
\begin{equation}
\mathbf{P_{lm}} = \begin{bmatrix}
    f       & 0 & 0 \\
    0      & f & 0
\end{bmatrix} [ \mathbf{R} * (\hat{\mathcal{M}}*w_\text{id}*w_\text{exp}) + \mathbf{t}]
\label{lm_eq}
\end{equation}
where $w_\text{exp}[1] = 1 - \sum_{i=2}^{47} w_\text{exp}[i]$ and $w_\text{exp}[i] \in [0,1], i = 2, \ldots, 47$. We use a unit quaternion $\mathbf{q} \in \mathbb{R}^{4 \times 1}$ \cite{pncc_new} to represent 3D rotation and convert it into rotation matrix for use in equation \ref{lm_eq}. Please note that, for retargeting purposes, we omit the learning of texture and lighting in the 3DMM.

\subsection{Multi-scale Representation Disentangling}
A straightforward way of holistically regressing all the 3DMM parameters together through a fully connected layer on top of one shared representation will not be optimal particularly for our problem where each group of parameters has strong semantic meanings. Intuitively speaking, head pose learning does not require detailed local face representations since it is fundamentally independent of skin texture and subtle facial expressions, which has also been observed in recent work on pose estimation \cite{tcdcn}. However, for identity learning, a combination of both local and global representations would be necessary to differentiate among different persons. For example, some persons have relatively small eyes but fat cheek while others have big eyes and thin cheek, so both the local features around the eyes and the overall face silhouette would be important to approximate the identity shape. Similarly, expression learning possibly requires even fine-grained granularity of different scales of representations. Single eye wink, mouth grin and big laugh clearly require three different levels of representations to differentiate them from other expressions.

\begin{figure}[t]
\centering
\includegraphics[width=0.32\linewidth]{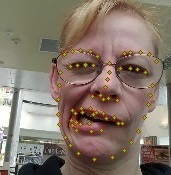}
\includegraphics[width=0.32\linewidth]{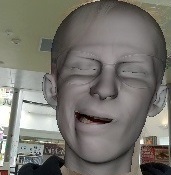}
\includegraphics[width=0.32\linewidth]{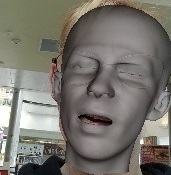}
\caption{\textbf{left}: landmark projection from both meshes are exactly the same, \textbf{mid}: mesh with maximum jaw left, \textbf{right}: mesh without jaw left, but larger roll angle}
\label{fig:decouple}
\end{figure}

\begin{figure}[t]
\centering
\includegraphics[width=0.19\linewidth]{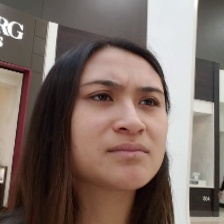}
\includegraphics[width=0.19\linewidth]{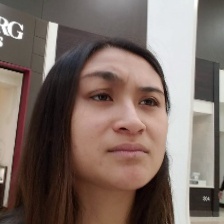}
\includegraphics[width=0.19\linewidth]{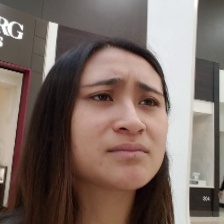}
\includegraphics[width=0.19\linewidth]{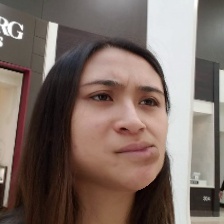}
\includegraphics[width=0.19\linewidth]{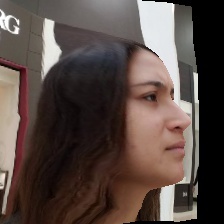}
\caption{Synthesis image for regularization}
\label{fig:synthesis}
\end{figure}

Another observation is, given the 2D landmarks of an image, there exist multiple combinations of 3DMM parameters that can minimize the 2D landmark loss. This ambiguity would cause additional challenges to the learning to favor the semantically more meaningful combinations. For examples, as shown in Fig. \ref{fig:decouple}, we can still minimize the 2D landmark loss by rotating the head and using different identity coefficients to accommodate the jaw left even without a strong jaw left expression coefficient. Motivated by both the multi-scale prior and the ambiguity nature of this problem, we designed a novel network structure that is specifically tailored for facial retargeting applications as illustrated in Fig. \ref{fig:teaser}, where pose is only learned through the final global features while expression learning depends on the concatenation of multi-layer representations.
\vspace{-10pt}
\paragraph{Disentangled Regularization} In addition to the above network design, we add a few regularization during the training to further enforce the disentangled representation learning. For example, for each face image, we can augment it by random translation/rotation perturbation to ask their resulting output to have the same identity and expression coefficients. Using image warping technique, we can re-edit the face image to slightly change the facial expression without hurting the pose and identity. Fig. \ref{fig:synthesis} shows a few such synthesized examples where their identity parameters need to be the same.

\subsection{Single Face Retargeting Network}\label{sec:sfn}
When the face bounding box is given, we can train a single face retargeting network to output 3DMM parameters for each cropped face image using the above proposed network structure. Fortunately, many public datasets \cite{300w, facewarehouse, LFWTech, 3ddfa} already provide bounding boxes along with 68 2D facial landmark points. To encourage disentangling, we fit 3DMM parameters for each cropped single face image using the optimization method of \cite{adrian3Dfacealignment} and treat them as ground truth for our network in addition to the landmarks. Although individual optimization may result in over-fitting and noisy ground truth, our network can intrinsically focus more on the global common patterns from the training data. To achieve this, we initially train with a large weight on the L1 loss with respect to the ground truth ($g$), and then gradually decay this weight to trust more on the 2D landmarks loss, as shown in the following loss function:
\begin{multline}
\tau * \Bigg\{\frac{1}{50}\sum_{i = 1}^{50}|w_{\text{id}_i} - {w}^g_{\text{id}_i}|
+\frac{1}{46}\sum_{i = 1}^{46}|w_{\text{exp}_i} - {w}^g_{\text{exp}_i}|    \\
+\frac{1}{4}\sum_{i = 1}^{4}|\mathbf{R}_i - \mathbf{R}^g_i|\Bigg\}
+ \sqrt{\frac{1}{68}\sum_{i = 1}^{68}(\mathbf{P}_{\text{lm}_i} - \mathbf{P}^g_{\text{lm}_i})^2}
\label{eq:sfn_loss}
\end{multline}
where $\tau$ denotes decay parameter with respect to epoch. We choose $\tau$ = 10/epoch across all experiments. Note that, although we drop the 3D translation and scale ground truth loss to allow 2D translation and scaling augmentation, the translation and scale parameters can still be learned by the 2D landmark loss.

\subsection{Joint Face Detection and Retargeting}
Our goal is to save computation cost by performing both face detection and 3DMM parameter estimation simultaneously instead of sequentially running a separate face detector and then single face retargeting network on each face separately. The network could potentially also benefit from the cross domain knowledge, especially for detection task, where introducing 3DMM gives the prior on how the face should look like in 3D space which complements the 2D features in separate face detection framework.

Inspired by YOLO \cite{yolo}, our joint network is designed to predict 3DMM parameters for each anchor point in additional to bounding box displacement and objectness. We divide the input image into $9 \times 9$ grid and predict a vector of length $4 + 1 + (50 + 46 + 4 + 3 + 1) =  109$ for a bounding box in each grid cell. Here 4 denotes 2D co-ordinates of the centroid, width and height of the face bounding box, 1 denotes the confidence score for the presence of a face in that cell and the rest denote the 3DMM parameters for the face in the cell. We also adopt the method of starting with 5 anchor boxes as bounding box priors. Our final loss function is the summation of equation \ref{eq:sfn_loss} across all grids and anchors, as shown in the following:
\begin{multline}
\tau * \Bigg\{\frac{1}{50}\sum_{j = 1}^{9^2} \sum_{k = 1}^{5}\sum_{i = 1}^{50}\mathbbm{1}_{ijk}|w_{\text{id}_{ijk}} - {w}^g_{\text{id}_{ijk}}|    \\
+\frac{1}{46}\sum_{j = 1}^{9^2} \sum_{k = 1}^{5}\sum_{i = 1}^{46}\mathbbm{1}_{ijk}|w_{\text{exp}_{ijk}} - {w}^g_{\text{exp}_{ijk}}|    \\
+\frac{1}{4}\sum_{j = 1}^{9^2} \sum_{k = 1}^{5}\sum_{i = 1}^{4}\mathbbm{1}_{ijk}|\mathbf{R}_{ijk} - \mathbf{R}^g_{ijk}|\Bigg\} \\ 
+ \sqrt{\frac{1}{68}\sum_{j = 1}^{9^2} \sum_{k = 1}^{5}\sum_{i = 1}^{68}\mathbbm{1}_{ijk}(\mathbf{P}_{\text{lm}_{ijk}} - \mathbf{P}^g_{\text{lm}_{ijk}})^2}\
\end{multline}
\label{eq:yolo_loss}
where $\mathbbm{1}_{ijk}$ denotes whether a $k$th bounding box predictor in cell $j$ contains a face. Since there are no publicly available multi-face datasets that provide both bounding box location and 3DMM parameters for each face, for proof-of-concept, we obtain the 3DMM ground truth by running our single face retargeting network on each face separately. The $x,y$ co-ordinates of the centroid and the width and height of a bounding box are calculated in the same manner as in \cite{yolo} and we use the same loss functions for these values.

\section{Experimental Setup}

\subsection{Datasets}
For single face retargeting, we combine multiple datasets to have a good training set for accurate prediction of each group of 3DMM parameters. 300W-LP contains many large poses and Facewarehouse is a rich dataset for expressions. The ground truth 68 2D landmarks provided by these datasets are used to obtain 3DMM ground truth by \cite{adrian3Dfacealignment}. LFW and AFLW2000-3D are used as test sets for static images and 300VW is used as test set for tracking on videos. For multiple face retargeting, AFW has ground truth bounding boxes, pose angles and 6 landmarks and is used as a test set for static images, while FDDB and WIDER only provide bounding box ground truth and are therefore used for training (WIDER test set is kept separate for testing). Music videos dataset is used to test our MFN performance on videos. We remove all images with more than 20 faces and also remove faces whose bounding box dimensions are \textless2\% of the image dimensions from both the training and test sets. This mainly includes faces in the background crowd with size less than 5$\times$5 pixels. The reason is that determining the facial expressions for such small faces is ambiguous even for human eyes and hence retargeting is not meaningful.
More dataset details are summarized in Table \ref{tab:dataset}. We use an 80-20 split of the training set for training and validation. To measure the performance of expression accuracy, we manually collect an expression test set by selecting those extreme expression images (Fig. \ref{fig:result4}). The number of images in each of the expression categories are: eye close: 185, eye wide: 70, brow raise: 124, brow anger: 100, mouth open: 81, jaw left/right: 136, lip roll: 64, smile: 105, kiss: 143, total: 1008 images.

\begin{table}[t]
\begin{tabular}{|l|l|c|c|}
\hline
\multicolumn{2}{|c|}{\textbf{Dataset}} & \multicolumn{1}{c|}{\textbf{\#images}} & \multicolumn{1}{c|}{\textbf{\#faces}} \\ \hline
\multirow{5}{*}{SFN}      & 300W-LP \cite{300w,3ddfa} & 61225 & 61225 \\ \cline{2-4} 
                          & FacewareHouse \cite{facewarehouse} & 5000  & 500 \\ \cline{2-4} 
                          & LFW \cite{LFWTech} & 12639  & 12639  \\ \cline{2-4} 
                          & AFLW2000-3D \cite{3ddfa} & 2000  & 2000 \\ \cline{2-4} 
                          & 300VW \cite{300VW} & 114 (videos)  & 218K\\
                          \hline
                          
\multirow{4}{*}{MFN}      & FDDB \cite{fddb} & 2845 & 5171 \\ \cline{2-4} 
                          & WIDER \cite{wider} & 11905 & 56525 \\ \cline{2-4} 
                          & AFW \cite{afwdatapaper} & 205 & 1000 \\ \cline{2-4} 
                          & Music videos \cite{musicvideo} & 8 (videos) & - \\
                          \hline
\end{tabular}
\caption {Number of images or videos and faces for each dataset used in training and testing of our networks.}
\label{tab:dataset}
\end{table}

\subsection{Evaluation Metrics}
We use 4 metrics: 1) average precision (AP) with different intersection-over-union thresholds as defined in \cite{yolov3} to evaluate our MFN performance for face detection, 2) normalized mean error (NME) defined as the Euclidean distance between the predicted and ground truth 2D landmarks averaged over 68 landmarks and normalized by the bounding box dimensions, 3) area under the curve (AUC) of the Cumulative Error Distribution curve for landmark error normalized by the diagonal distance of ground truth bounding box \cite{Jointalignment}, and 4) expression metric defined as the mean absolute distance between the predicted expression parameters with respect to the ground truth, which is 1 in our case following the practice of \cite{facewarehouse}. 

\subsection{Implementation Details}

\subsubsection{Training Configuration}
 Our networks are implemented in Keras \cite{chollet2015keras} with Tensorflow backend and trained using Adam optimizer with batch size 32. The initial learning rate ($10^{-3}$ for SFN and $10^{-4}$ for MFN) is decreased by 10 times (up to $10^{-6}$) when the validation loss does not change over 5 consecutive epochs. Training takes about a day on a Nvidia GTX 1080 for each network. For data augmentation, we use random scaling in the range [0.8,1.2], random translation of 0-10\%, color jitter and in-plane rotation. These augmentation techniques improve the performance of SFN and also help in generating more accurate ground truth for individual faces for MFN.

\subsubsection{Single Face Retargeting Architecture}
Our network takes 128x128 resized image as input. In the first layer, we use a $7 \times 7$ convolution layer with 64 filters and stride 2 followed by a $2\times2$ maxpooling layer to capture the fine details in the image. The following layers are made up of Fire Modules(FM) of SqueezeNet \cite{SqueezeNet} (with 16 and 64 filters in squeeze and expand layers respectively) and squeeze-and-excite modules(SE) of \cite{SEnet} in order to compress the model size and reduce the model execution time without compromising the accuracy. At the end of network, we use a global average pooling layer followed by fully connected (FC) layers to generate the parameters. The penultimate FC layers each has 64 units with ReLU activation and sigmoid activation is used at the end of the last expression branch's FC layer  to restrict the values between 0 and 1. To realize the multiscale prior and the disentangled learning, we concatenate the features at different scales and form separate branches for each group of parameters. The extra branches are built with the same blocks as the main branch, but we reduce the channel size by half to restrict the extra computation cost.

\subsubsection{Joint Detection and Retargeting Architecture}
Our joint detection and retargeting network architecture is similar to Tiny DarkNet \cite{yolo} with the final layer changed to predict a tensor of size $9\times 9 \times 5 \times 109$. However, since we only have one object class (face) in our problem, we reduce the number of filters in each layer to a quarter of their original values. For multi-scale version, we change the input image size to $288 \times 288$ and extend the multi-scale backbone for single face retargeting by changing the output of each branch to accommodate grid output (Fig. \ref{fig:teaser}). The pose branch outputs change from 4 ($R$) + 3 ($T$) + 1 ($f$) = 8 to $9\times 9 \times 5 \times 8$. The expression branch outputs change from 46 to $9\times 9 \times 5 \times 46$, and identity branch outputs change from 50 to $9\times 9 \times 5 \times 50$. One extra branch is also added to output objectness and bounding box location ($9\times 9 \times 5 \times (4+1)$). In total, multi-scale version outputs the same dimension as single-scale, but the output channels are split with respect to each type of branch.

\begin{figure}[t]
\centering
\begin{subfigure}[t]{1.0\linewidth}
\centering
\begin{subfigure}[t]{0.19\textwidth}
\centering
\includegraphics[width = 0.95\textwidth, height = 0.065\textheight]{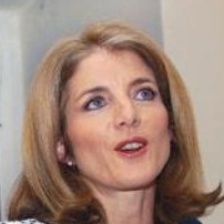}
\end{subfigure}
\begin{subfigure}[t]{0.19\textwidth}
\centering
\includegraphics[width = 0.95\textwidth, height = 0.065\textheight]{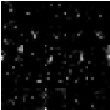}
\end{subfigure}
\begin{subfigure}[t]{0.19\textwidth}
\centering
\includegraphics[width = 1.0\textwidth, height = 0.065\textheight]{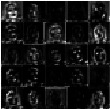}
\end{subfigure}
\begin{subfigure}[t]{0.19\textwidth}
\centering
\includegraphics[width = 1.0\textwidth, height = 0.065\textheight]{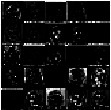}
\end{subfigure}
\begin{subfigure}[t]{0.19\textwidth}
\centering
\includegraphics[width = 1.0\textwidth, height = 0.065\textheight]{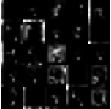}
\end{subfigure}
\end{subfigure}
\begin{subfigure}[t]{1.0\linewidth}
\centering
\begin{subfigure}[t]{0.19\textwidth}
\centering
\includegraphics[width = 0.95\textwidth, height = 0.065\textheight]{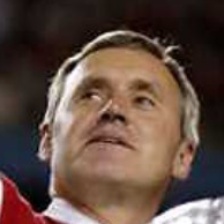}
\end{subfigure}
\begin{subfigure}[t]{0.19\textwidth}
\centering
\includegraphics[width = 0.95\textwidth, height = 0.065\textheight]{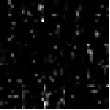}
\end{subfigure}
\begin{subfigure}[t]{0.19\textwidth}
\centering
\includegraphics[width = 1.0\textwidth, height = 0.065\textheight]{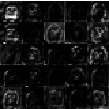}
\end{subfigure}
\begin{subfigure}[t]{0.19\textwidth}
\centering
\includegraphics[width = 1.0\textwidth, height = 0.065\textheight]{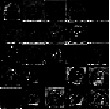}
\end{subfigure}
\begin{subfigure}[t]{0.19\textwidth}
\centering
\includegraphics[width = 1.0\textwidth, height = 0.065\textheight]{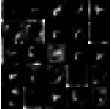}
\end{subfigure}
\end{subfigure}
\caption{Visualization of learned features. From left to right in each row: input image, features for single scale SFN, features for expression branch of multi-scale SFN, features for identity branch of multi-scale SFN, features for pose branch of multi-scale SFN.}
\label{fig:layeroutputs}
\end{figure}

\section{Results}
\subsection{Importance of Multi-Scale Representation}
\begin{figure}[t]
\centering
\begin{subfigure}[t]{1.0\linewidth}
\centering
\begin{subfigure}[t]{0.24\textwidth}
\centering
\includegraphics[width = 1.0\textwidth, height = 0.08\textheight]{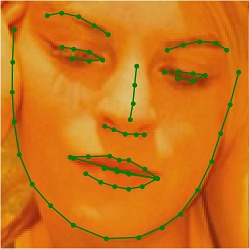}
\end{subfigure}
\begin{subfigure}[t]{0.24\textwidth}
\centering
\includegraphics[width = 1.0\textwidth, height = 0.08\textheight]{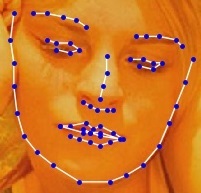}
\end{subfigure}
\begin{subfigure}[t]{0.24\textwidth}
\centering
\includegraphics[width = 1.0\textwidth, height = 0.08\textheight]{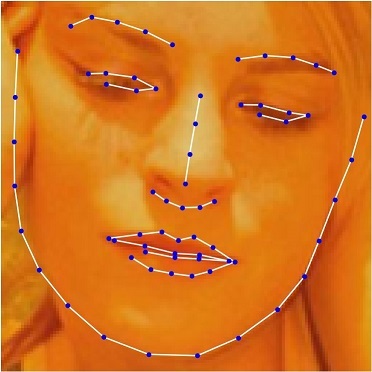}
\end{subfigure}
\begin{subfigure}[t]{0.24\textwidth}
\centering
\includegraphics[width = 1.0\textwidth, height = 0.08\textheight]{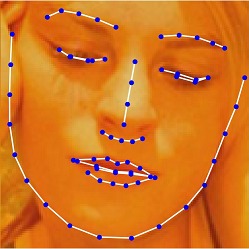}
\end{subfigure}
\end{subfigure}

\begin{subfigure}[t]{1.0\linewidth}
\centering
\begin{subfigure}[t]{0.24\textwidth}
\centering
\includegraphics[width = 1.0\textwidth, height = 0.08\textheight]{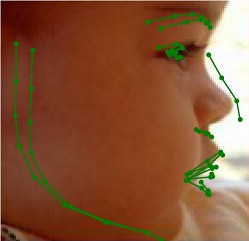}
\end{subfigure}
\begin{subfigure}[t]{0.24\textwidth}
\centering
\includegraphics[width = 1.0\textwidth, height = 0.08\textheight]{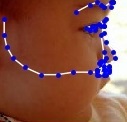}
\end{subfigure}
\begin{subfigure}[t]{0.24\textwidth}
\centering
\includegraphics[width = 1.0\textwidth, height = 0.08\textheight]{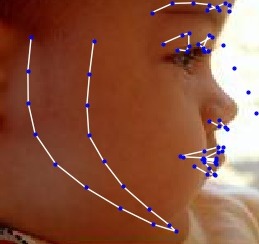}
\end{subfigure}
\begin{subfigure}[t]{0.24\textwidth}
\centering
\includegraphics[width = 1.0\textwidth, height = 0.08\textheight]{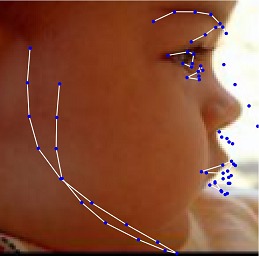}
\end{subfigure}
\end{subfigure}

\begin{subfigure}[t]{1.0\linewidth}
\centering
\begin{subfigure}[t]{0.24\textwidth}
\centering
\includegraphics[width = 1.0\textwidth, height = 0.08\textheight]{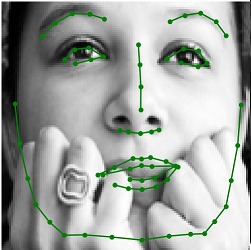}
\end{subfigure}
\begin{subfigure}[t]{0.24\textwidth}
\centering
\includegraphics[width = 1.0\textwidth, height = 0.08\textheight]{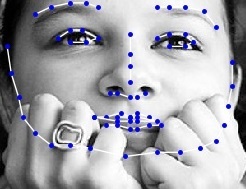}
\end{subfigure}
\begin{subfigure}[t]{0.24\textwidth}
\centering
\includegraphics[width = 1.0\textwidth, height = 0.08\textheight]{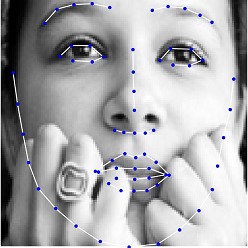}e
\end{subfigure}
\begin{subfigure}[t]{0.24\textwidth}
\centering
\includegraphics[width = 1.0\textwidth, height = 0.08\textheight]{images/stage2res6.jpg}
\end{subfigure}
\end{subfigure}
\caption{2D Face Alignment results for AFLW2000-3D. First column: original image with ground truth landmarks; Second column: results using \cite{adrian3Dfacealignment}; Third column: our single scale SFN; Last column: our multi-scale SFN.}
\label{fig:howfarcomparison}
\end{figure}

Our multi-scale network design reduces the load on the network to learn complex features by allowing the network to concentrate on different image features to learn different parameters unlike the single scale design. In Fig. \ref{fig:layeroutputs}, we see that single scale network learns generic facial features that combines the representations for identity, expression and pose. On the other hand, multi-scale network learns different levels of representation (pixel-level detailed features for expression, region level features for identity and global aggregate features for pose). We have randomly chosen only 25 filter outputs at level 3 of our SFN for clearer visualization. Table \ref{table:full} shows that our multi-scale design not only reduces NME for single face images using SFN but also improves the performance of MFN in terms of both NME (by generating a better weakly supervised ground truth) and AP for detection. Clearly, different feature representations are crucial to accurately learn different groups of parameters. By reducing the network load, this design also allows model compression so that multi-scale networks can be of comparable size with respect to single scale networks while having better accuracy. Fig. \ref{fig:howfarcomparison} shows that the multi-scale design predicts more accurate expression parameters (first row has correct landmarks for closed eyes) and identity parameters (second row has correct landmarks that fit the face shape) while being robust to large poses (second row), illumination (first row) and occlusion (third row).

\begin{table}[t]
\small
\begin{tabular}{|l|c|c|c|c|}
\hline
\multicolumn{1}{|c|}{\multirow{3}{*}{Model}} & \multicolumn{4}{c|}{Evaluation} \\ 
\cline{2-5} \multicolumn{1}{|c|}{}  &  \multirow{2}{*}{\begin{tabular}[c]{@{}c@{}}NME\\ (\%)\end{tabular}} & \multicolumn{3}{c|}{Multi Face}  \\ 
\cline{3-5} \multicolumn{1}{|c|}{}  &      & \multicolumn{1}{l|}{AP} & \multicolumn{1}{l|}{AP50} & \multicolumn{1}{l|}{AP75} \\ \hline
(1) MFN (detection only)  & - & 92.1 & 99.2 & 94.3\\ \hline
(2) Single scale SFN & 2.16 & - & - & -\\ \hline
(3) Multi-scale SFN  & 1.91 & - & - & - \\ \hline
(4) SS-MFN + GT from (2) & 2.89 & 97.5 & 99.8 & 98.1  \\ \hline
(5) SS-MFN + GT from (3)  & 2.65 & 98.2 & 100 & 98.9  \\ \hline
(6) MS-MFN + GT from (3)  & 2.23 & 98.8 & 100 & 99.3\\ \hline
\end{tabular}
\caption{Quantitative evaluation of our SFN and MFN. SS-MFN and MS-MFN denote single scale and multi-scale MFN respectively. NME values are calculated for LFW (single faces) and AP values are calculated for AFW.}
\label{table:full}
\end{table}

\subsection{Comparison with 2D Alignment Methods}
Even though we aim to predict the 3DMM parameters for retargeting applications, our model can naturally serve the purpose for 2D face alignment (via 3D). Therefore, we can evaluate our model from the performance of 2D alignment perspective. Table \ref{table:aflw_nme} compares the performance of our single scale and multi-scale SFN with state-of-the-art 2D face alignment methods (compared under the same settings). As can be seen, our model achieves much smaller errors compared to most of the methods that are dedicated for precise landmark localization. While PRN \cite{prn} has lower NME, its network size is 80 times bigger than ours and takes 9.8ms on GPU compared to $<$1ms required by our network. In addition to evaluations on static images, we also measure the face tracking performance in a video using our SFN. We set the bounding box of the current frame using the boundaries of the 2D landmarks detected in the previous frame and perform retargeting on a frame-by-frame basis. Table \ref{table:300vw_auc} compares the AUC values on 300VW dataset for three scenarios categorized by the dataset (compared under the same settings). Our method performs significantly better than other methods (about 9\% improvement over the second best method for Scenario 3) with negligible failure rate because extensive data augmentation helps our tracking algorithm to quickly recover from failures.

\begin{table}[t]
\small
    \centering
    \begin{tabular}{|c||c|c|c|c|}
    \hline
     Method  &  [0\degree,30\degree] & [30\degree,60\degree] & [60\degree,90\degree] & Mean\\
     \hline
     SDM \cite{sdm}  &  3.67 & 4.94 & 9.67 & 6.12 \\
     \hline
     3DDFA \cite{3ddfa} & 3.78 & 4.54 & 7.93 & 5.42\\
     \hline
     3DDFA2 \cite{3ddfa} & 3.43 & 4.24 & 7.17 & 4.94\\
     \hline
     Yu et al. \cite{Yu2017LearningDF} & 3.62 & 6.06 & 9.56 & 6.41\\
     \hline
     3DSTN \cite{FasterTR} & 3.15 & 4.33 & 5.98 & 4.49\\
     \hline
     DFF \cite{dff}  & 3.20 & 4.68 & 6.28 & 4.72 \\
     \hline
     PRN \cite{prn} & 2.75 & 3.51 & 4.61 & 3.62\\
     \hline
     SS-SFN (ours) & \textbf{3.09} & \textbf{4.27} & \textbf{5.59} & \textbf{4.31}\\
     \hline
     MS-SFN (ours) & \textbf{2.91} & \textbf{3.83} & \textbf{4.94} & \textbf{3.89}\\
     \hline
    \end{tabular}
    \caption{Comparison of NME(\%) for 68 landmarks for AFLW2000-3D (divided into 3 groups based on yaw angles). 3DDFA2 refers to 3DDFA+SDM \cite{3ddfa}.}
    \label{table:aflw_nme}
\end{table}

\begin{table}[t]
\small
    \centering
    \begin{tabular}{|c||c|c|c|}
    \hline
     Method & Scenario 1 & Scenario 2 & Scenario 3\\
     \hline
      Yang et al. \cite{yangetal} & 0.791 & 0.788 & 0.710\\
      \hline
      Xiao et al. \cite{xiaoetal} & 0.760 & 0.782 & 0.695 \\
      \hline
      CFSS \cite{cfss} & 0.784 & 0.783 & 0.713 \\
      \hline
      MTCNN \cite{MTCNN} & 0.748 & 0.760 & 0.726\\
      \hline
      MHM \cite{Jointalignment} & 0.847 & 0.838 & 0.769\\
      \hline
      MS-SFN (ours) & \textbf{0.901} & \textbf{0.884} & \textbf{0.842}\\
      \hline
    \end{tabular}
    \caption{Landmark localization performance of our method on videos in comparison to state-of-the-art face tracking methods. The values are reported in terms of Area under the Curve (AUC) for Cumulative Error Distribution of the 2D landmark error for 300VW test set.}
    \label{table:300vw_auc}
\end{table}

\begin{figure*}[h!]
\centering
\begin{subfigure}[t]{0.16\linewidth}
\includegraphics[width = 1.0\textwidth, height = 0.1\textheight]{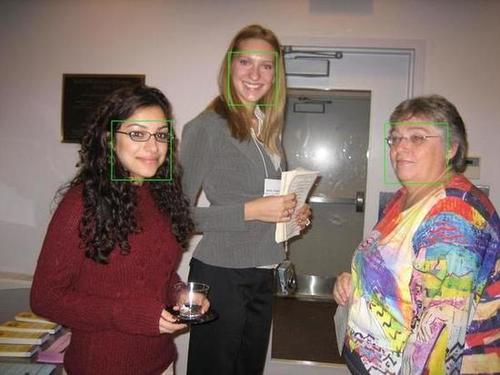}
\end{subfigure}
\begin{subfigure}[t]{0.16\linewidth}
\includegraphics[width = 1.0\textwidth, height = 0.1\textheight]{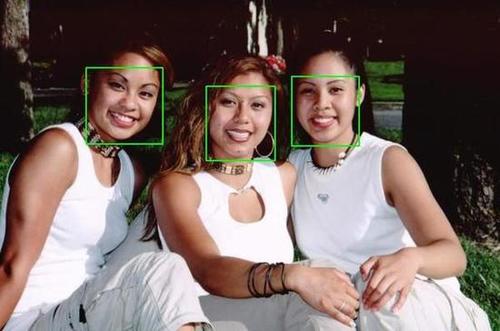}
\end{subfigure}
\begin{subfigure}[t]{0.16\linewidth}
\includegraphics[width = 1.0\textwidth, height = 0.1\textheight]{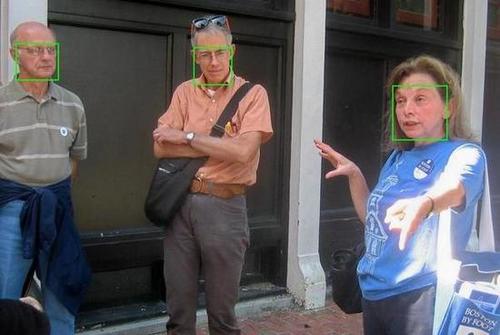}
\end{subfigure}
\begin{subfigure}[t]{0.16\linewidth}
\includegraphics[width = 1.0\textwidth, height = 0.1\textheight]{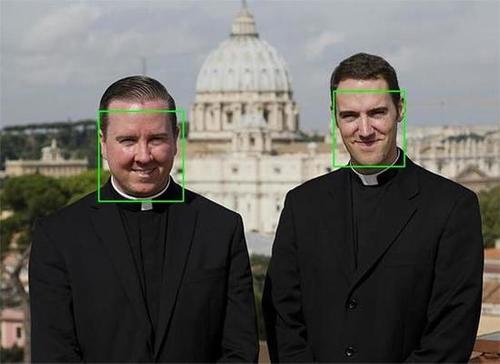}
\end{subfigure}
\begin{subfigure}[t]{0.16\linewidth}
\includegraphics[width = 1.0\textwidth, height = 0.1\textheight]{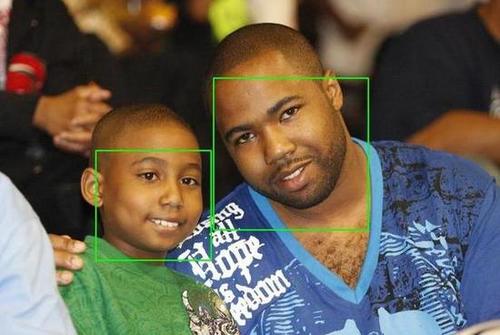}
\end{subfigure}
\begin{subfigure}[t]{0.16\linewidth}
\includegraphics[width = 1.0\textwidth, height = 0.1\textheight]{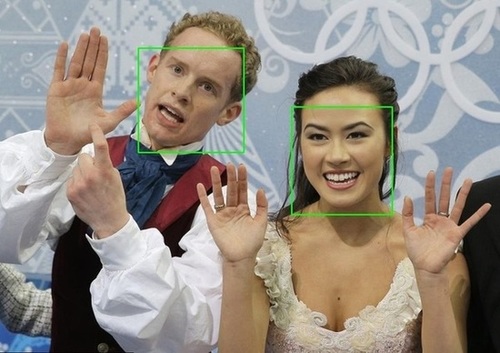}
\end{subfigure}
\begin{subfigure}[t]{0.16\linewidth}
\includegraphics[width = 1.0\textwidth, height = 0.1\textheight]{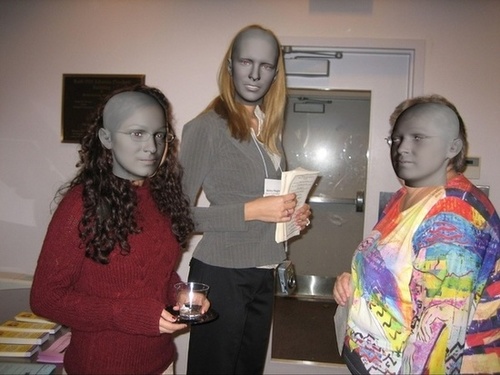}
\end{subfigure}
\begin{subfigure}[t]{0.16\linewidth}
\includegraphics[width = 1.0\textwidth, height = 0.1\textheight]{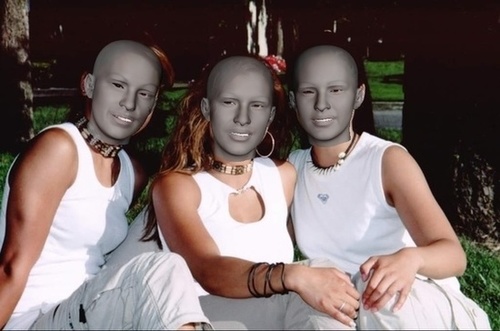}
\end{subfigure}
\begin{subfigure}[t]{0.16\linewidth}
\includegraphics[width = 1.0\textwidth, height = 0.1\textheight]{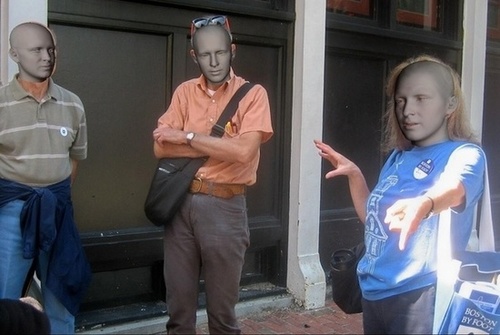}
\end{subfigure}
\begin{subfigure}[t]{0.16\linewidth}
\includegraphics[width = 1.0\textwidth, height = 0.1\textheight]{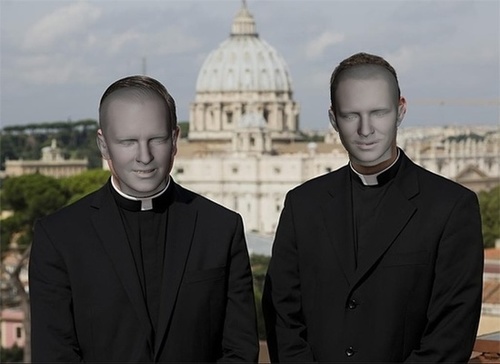}
\end{subfigure}
\begin{subfigure}[t]{0.16\linewidth}
\includegraphics[width = 1.0\textwidth, height = 0.1\textheight]{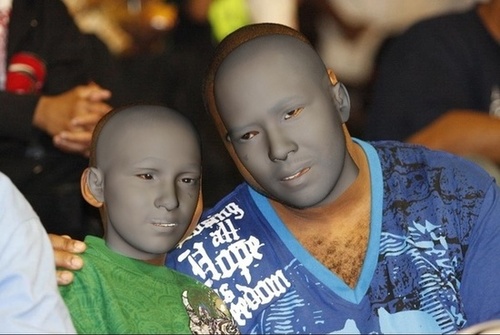}
\end{subfigure}
\begin{subfigure}[t]{0.16\linewidth}
\includegraphics[width = 1.0\textwidth, height = 0.1\textheight]{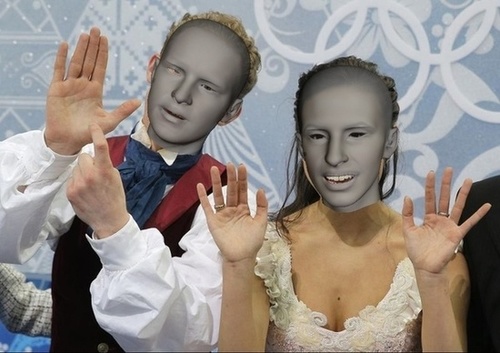}
\end{subfigure}
\caption{Testing results of our joint detection and retargeting model. 
Column \textbf{1-3}: Sampled from AFW; Column \textbf{4-6}: Sampled from WIDER. We show both the predicted bounding boxes and the 3D meshes constructed from 3DMM parameters. }
\label{fig:result1}
\end{figure*}

\begin{table*}[t]
\begin{tabular}{|l|l|l|l|l|l|l|l|l|l|l|}
\hline
\multicolumn{1}{|c|}{Model} & \multicolumn{1}{c|}{\begin{tabular}[c]{@{}c@{}}Eye\\ Close\end{tabular}} & \multicolumn{1}{c|}{\begin{tabular}[c]{@{}c@{}}Eye\\ Wide\end{tabular}} & \multicolumn{1}{c|}{\begin{tabular}[c]{@{}c@{}}Brow\\ Raise\end{tabular}} & \multicolumn{1}{c|}{\begin{tabular}[c]{@{}c@{}}Brow\\ Anger\end{tabular}} & \multicolumn{1}{c|}{\begin{tabular}[c]{@{}c@{}}Mouth\\ Open\end{tabular}} & \multicolumn{1}{c|}{\begin{tabular}[c]{@{}c@{}}Jaw\\ L/R\end{tabular}} & \multicolumn{1}{c|}{\begin{tabular}[c]{@{}c@{}}Lip\\ Roll\end{tabular}} & Smile & Kiss & \multicolumn{1}{c|}{Avg} \\ \hline
(1) Single scale SFN & 0.082   & 0.265   & 0.36   & 0.451   & 0.373   & 0.331   & 0.359   & 0.223 & 0.299 & 0.305 \\ \hline
(2) Multi-scale SFN & 0.016   & 0.257   & 0.216   & 0.381   & 0.334   & 0.131   & 0.204   & 0.245 & 0.277 & 0.229 \\ \hline
(3) MS-MFN + GT from (2) & 0.117   & 0.407   & 0.284   & 0.405   & 0.284   & 0.173   & 0.325   & 0.248 & 0.349 & 0.288 \\ \hline
\end{tabular}
\caption{Quantitative evaluation of expression accuracy (measured by the expression metric) on our expression test set. Lower error means the model performs better for extreme expressions when required.}
\label{table:exp}
\end{table*}

\begin{figure*}[h!]
\centering
\begin{subfigure}[t]{0.118\linewidth}
\includegraphics[width = 1.0\textwidth, height = 0.09\textheight]{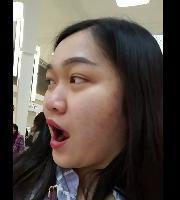}
\end{subfigure}
\begin{subfigure}[t]{0.118\linewidth}
\includegraphics[width = 1.0\textwidth, height = 0.09\textheight]{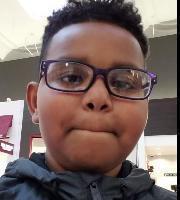}
\end{subfigure}
\begin{subfigure}[t]{0.118\linewidth}
\includegraphics[width = 1.0\textwidth, height = 0.09\textheight]{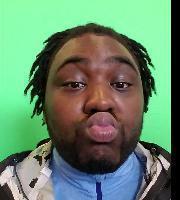}
\end{subfigure}
\begin{subfigure}[t]{0.118\linewidth}
\includegraphics[width = 1.0\textwidth, height = 0.09\textheight]{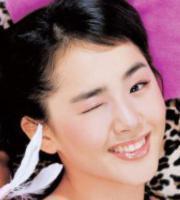}
\end{subfigure}
\begin{subfigure}[t]{0.118\linewidth}
\includegraphics[width = 1.0\textwidth, height = 0.09\textheight]{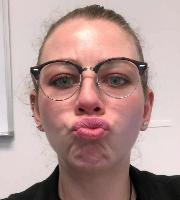}
\end{subfigure}
\begin{subfigure}[t]{0.118\linewidth}
\includegraphics[width = 1.0\textwidth, height = 0.09\textheight]{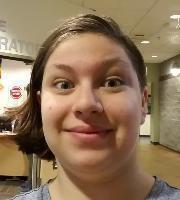}
\end{subfigure}
\begin{subfigure}[t]{0.118\linewidth}
\includegraphics[width = 1.0\textwidth, height = 0.09\textheight]{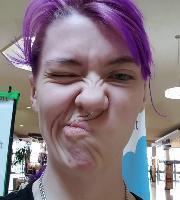}
\end{subfigure}
\begin{subfigure}[t]{0.118\linewidth}
\includegraphics[width = 1.0\textwidth, height = 0.09\textheight]{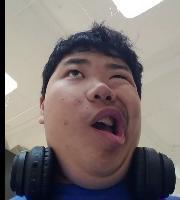}
\end{subfigure}

\begin{subfigure}[t]{0.118\linewidth}
\includegraphics[width = 1.0\textwidth, height = 0.09\textheight]{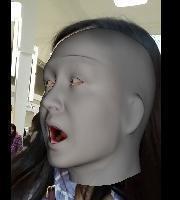}
\end{subfigure}
\begin{subfigure}[t]{0.118\linewidth}
\includegraphics[width = 1.0\textwidth, height = 0.09\textheight]{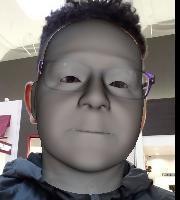}
\end{subfigure}
\begin{subfigure}[t]{0.118\linewidth}
\includegraphics[width = 1.0\textwidth, height = 0.09\textheight]{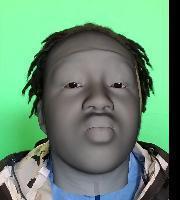}
\end{subfigure}
\begin{subfigure}[t]{0.118\linewidth}
\includegraphics[width = 1.0\textwidth, height = 0.09\textheight]{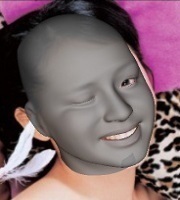}
\end{subfigure}
\begin{subfigure}[t]{0.118\linewidth}
\includegraphics[width = 1.0\textwidth, height = 0.09\textheight]{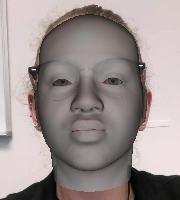}
\end{subfigure}
\begin{subfigure}[t]{0.118\linewidth}
\includegraphics[width = 1.0\textwidth, height = 0.09\textheight]{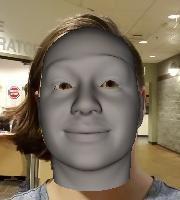}
\end{subfigure}
\begin{subfigure}[t]{0.118\linewidth}
\includegraphics[width = 1.0\textwidth, height = 0.09\textheight]{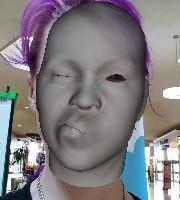}
\end{subfigure}
\begin{subfigure}[t]{0.118\linewidth}
\includegraphics[width = 1.0\textwidth, height = 0.09\textheight]{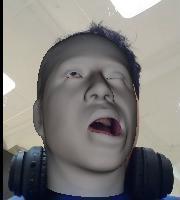}
\end{subfigure}
\caption{Results by applying MS-SFN on our expression test set.}
\label{fig:result4}
\end{figure*}

\begin{figure*}[h!]
\centering
\includegraphics[width=1.0\linewidth]{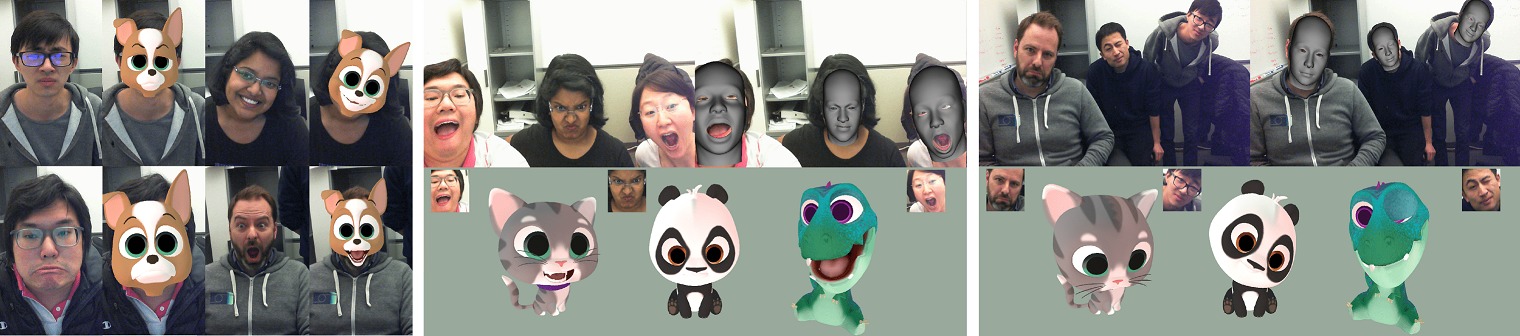}
\caption{Retargeting from face(s) to 3D character(s).}
\label{fig:retargeting}
\end{figure*}

\subsection{Importance of Joint Training}
Joint regression of both face bounding box locations and 3DMM parameters forces the network to learn exclusive facial features that characterize face shape, expression and pose in addition to differentiating face regions from the background. This helps in more precise face detection in-the-wild by leveraging both 2D information from bounding boxes and 3D information from 3DMM parameters. Table \ref{table:full} shows that average precision (AP) is improved by a large margin with joint training compared to when the same network is trained to only regress bounding box locations. The retargeting accuracy for MFN is also comparable to that of SFN and the slight decrease in NME is because of training MFN on multi-face images and testing on single face images. Nevertheless, we observe improved performance in terms of both NME and AP by using better ground truth generated by multi-scale model. Our detection accuracy (mAP: 98.8\%) outperforms Hyperface \cite{hyperface} (mAP: 97.9\%) and Faceness-Net \cite{facenessnet} (mAP: 97.2\%) on the entire AFW dataset when compared under the same settings. Results of our MFN on multi-face images are illustrated in Fig. \ref{fig:result1}.

\subsection{Evaluation of Expressions}
Our expression evaluation results in Table \ref{table:exp} emphasizes the improvement of multi-scale design on SFN. MS-MFN performs much better than SS-SFN for all expressions except the eye expressions. This is because eye patches are usually small compared to the entire image for MFN whereas they are zoomed in on cropped images for SFN. Attention network for emphasizing small eyes region could be a future work for our MFN. However, MS-MFN shows less accuracy compared to MS-SFN because it is being tested on single face images while being trained on multi-face images. For the multi-person test set images, we found similar visual results by applying MS-MFN on the whole image and by applying MS-SFN on each face individually cropped from the image. This is expected because MFN is trained with ground truth from SFN. The performance of MS-SFN on our expression test set is shown in Fig. \ref{fig:result4}. We also conducted live performance capture experiments to evaluate the efficiency our system in retargeting facial motion from one or more faces to one or more 3D characters. Fig. \ref{fig:retargeting} shows some screenshots recorded during the experiments.

\subsection{Computational Complexity}
Excluding the IO time, SFN can run at 15ms/frame on Google Pixel 2 (assuming single face and excluding face detector runtime). Face detection with our compressed detector model is 34ms, so separate face detection and retargeting requires 49ms for 1 face, 109ms for 5 faces and 184ms for 10 faces. On the other hand, our proposed MFN performs joint face detection and retargeting at 39ms on any number of faces. The model sizes for compressed face detector is 11.5MB and SFN is 2MB, so the combination is 13.5MB, while our MFN is only 13MB. Hence our joint network reduces both memory requirement and execution time.

\section{Conclusion}
We propose a lightweight multitask learning network for joint face detection and facial motion retargeting on mobile devices in real time. The lack of 3DMM training data for multiple faces is tackled by generating weakly supervised ground truth from a network trained on images with single faces. We carefully design the network structure and regularization to enforce disentangled representation learning inspired by key observations. Extensive results have demonstrated the effectiveness of our design.

\textbf{Acknowledgements} We thank the anonymous reviewers for their constructive feedback, Muscle Wu, Xiang Yan, Zeyu Chen and Pai Zhang for helping, and Linda Shapiro, Alex Colburn and Barbara Mones for valuable discussions.

\newpage

\section{Appendix}
Our goal is to perform live facial motion retargeting from multiple faces in a frame to 3D characters on mobile devices. We propose a lightweight multi-scale network architecture to disentangle the 3DMM parameters so that only the expression and pose parameters can be seamlessly transferred to any 3D character. Furthermore, we avoid the performance overhead of running a separate face detector (as in \cite{DDEregression}) by integrating the face detection task with the parameter estimation task.

\subsection{Network Topology}
\begin{figure}[t]
\centering
\includegraphics[width=1.0\linewidth]{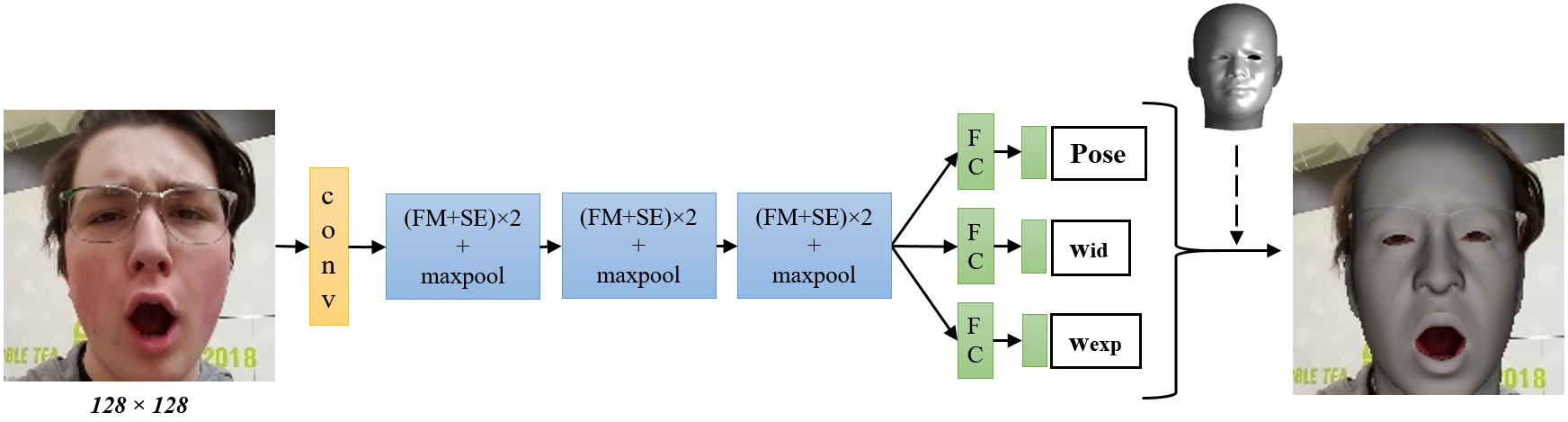}
\caption{Our single scale single face retargeting network}
\label{fig:ss-sfn}
\end{figure}

The architecture of our single scale single face retargeting network is shown in Fig. \ref{fig:ss-sfn}. The details of each block are given in our paper. As can be seen, the resolution (scale) of the image feature maps is reduced by 2 after every block, resulting in a $8 \times 8$ feature map at the end before global average pooling. The network then needs to learn pose, expression and identity parameters from the same $8\times8$ feature map, hence the term single scale.
In retargeting applications, it is necessary to disentangle expression and rotation parameters from the rest (identity, translation and scale) and accurately predict each group of parameters. Besides, we have argued how low-level features are important for expression and high-level features are important for head pose. Keeping these two requirements in mind, we designed our multi-scale single face retargeting network to learn different groups of parameters from separate branches that represent image features at different scales.

\subsection{Multi-face Retargeting Network Outputs}
As mentioned in our paper, the multi-face retargeting network divides the input image into $9\times9$ grid and predicts 5 bounding boxes for each grid cell. Each bounding box $b$ has the following co-ordinates: $t_x$, $t_y$, $t_w$, $t_h$, $t_o$ and $t_{v_{1-104}}$. The final outputs ($b_x, b_y$ - $x,y$ co-ordinates of the box centroid, $b_w, b_h$ - width and height of the box, $b_o$ - objectness score, $b_{\text{id}}, b_{\text{exp}}, b_{\textbf{R}}, b_{\textbf{t}}, b_{\text{f}}$ - 3DMM parameters and $b_{\text{lm}}$ - corresponding 2D landmarks) are then given by:
\begin{equation*}
    b_x = \sigma (t_x) + c_x;\; b_y = \sigma (t_y) + c_y
\end{equation*}
\begin{equation*}
    b_w = p_w * e^{t_w}; \; b_h = p_h * e^{t_h}
\end{equation*}
\begin{equation*}
    b_o = Pr(\text{face}) * IOU(b,\text{face}) = \sigma (t_o)
\end{equation*}
\begin{equation*}
    b_{\text{id}} = t_{v_{1-50}}; \; b_{\text{exp}} = \sigma (t_{v_{51-97}})
\end{equation*}
\begin{equation*}
    b_{\textbf{R}} = t_{v_{98-101}}; \; b_{\textbf{t}} = t_{v_{102-104}}; \; b_{\text{f}} = \sigma (t_{v_{105}})
\end{equation*}
\begin{equation*}
    b_{\text{lm}_x} = b_x + b_w * b_{\Hat{\text{lm}}_x}; \; b_{\text{lm}_y} = b_y + b_h * b_{\Hat{\text{lm}}_y}
\end{equation*}
where $\sigma$ denotes sigmoid function, $(c_x,c_y)$ is the offset of the grid cell containing $b$ from the top left corner of the image, $(p_w,p_h)$ are the dimensions of the bounding box prior (anchor box), $b_{\Hat{\text{lm}}}$ are the initial landmarks obtained using and IOU denotes intersection over union. As evident from the equation, the landmark loss puts additional constraints on the bounding box locations and dimensions. This contributes to our observation that joint face detection and regression of 3DMM parameters improves the accuracy of face detection compared to simple face detection.

\begin{figure}[t]
\centering
\includegraphics[width=0.9\linewidth]{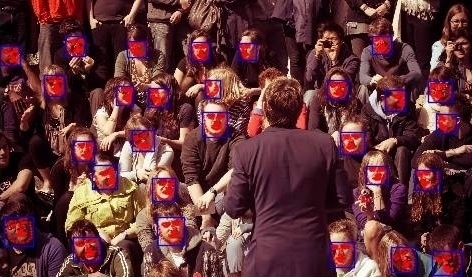}
\caption{Network output for an image with multiple small faces from the AFW dataset. The blue rectangles denote the predicted bounding boxes and the red points denote the 2D landmarks (for better viewing) projected from the predicted 3DMM parameters.}
\label{fig:facedetection}
\end{figure}

\begin{figure*}[h!]
\centering
\begin{subfigure}[t]{0.118\linewidth}
\includegraphics[width = 1.0\textwidth, height = 0.09\textheight]{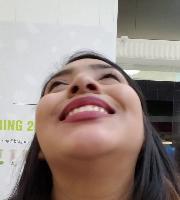}
\end{subfigure}
\begin{subfigure}[t]{0.118\linewidth}
\includegraphics[width = 1.0\textwidth, height = 0.09\textheight]{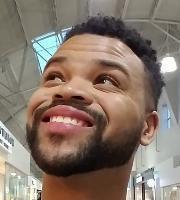}
\end{subfigure}
\begin{subfigure}[t]{0.118\linewidth}
\includegraphics[width = 1.0\textwidth, height = 0.09\textheight]{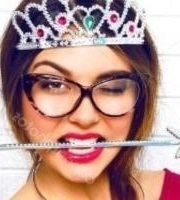}
\end{subfigure}
\begin{subfigure}[t]{0.118\linewidth}
\includegraphics[width = 1.0\textwidth, height = 0.09\textheight]{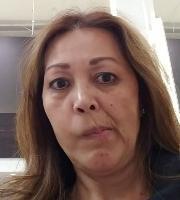}
\end{subfigure}
\begin{subfigure}[t]{0.118\linewidth}
\includegraphics[width = 1.0\textwidth, height = 0.09\textheight]{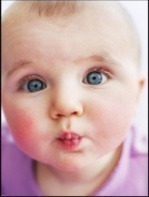}
\end{subfigure}
\begin{subfigure}[t]{0.118\linewidth}
\includegraphics[width = 1.0\textwidth, height = 0.09\textheight]{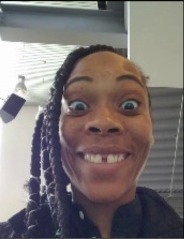}
\end{subfigure}
\begin{subfigure}[t]{0.118\linewidth}
\includegraphics[width = 1.0\textwidth, height = 0.09\textheight]{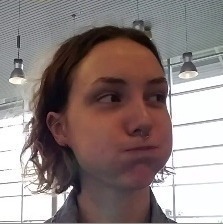}
\end{subfigure}
\begin{subfigure}[t]{0.118\linewidth}
\includegraphics[width = 1.0\textwidth, height = 0.09\textheight]{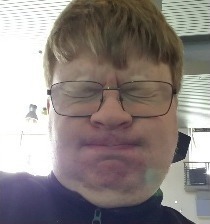}
\end{subfigure}

\begin{subfigure}[t]{0.118\linewidth}
\includegraphics[width = 1.0\textwidth, height = 0.09\textheight]{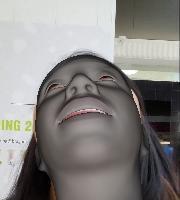}
\end{subfigure}
\begin{subfigure}[t]{0.118\linewidth}
\includegraphics[width = 1.0\textwidth, height = 0.09\textheight]{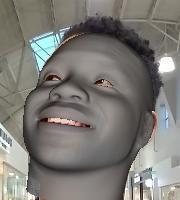}
\end{subfigure}
\begin{subfigure}[t]{0.118\linewidth}
\includegraphics[width = 1.0\textwidth, height = 0.09\textheight]{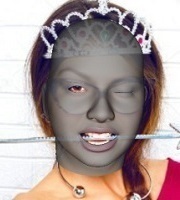}
\end{subfigure}
\begin{subfigure}[t]{0.118\linewidth}
\includegraphics[width = 1.0\textwidth, height = 0.09\textheight]{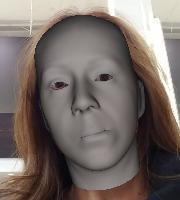}
\end{subfigure}
\begin{subfigure}[t]{0.118\linewidth}
\includegraphics[width = 1.0\textwidth, height = 0.09\textheight]{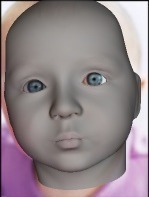}
\end{subfigure}
\begin{subfigure}[t]{0.118\linewidth}
\includegraphics[width = 1.0\textwidth, height = 0.09\textheight]{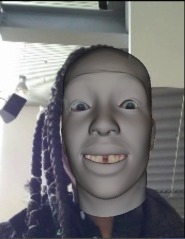}
\end{subfigure}
\begin{subfigure}[t]{0.118\linewidth}
\includegraphics[width = 1.0\textwidth, height = 0.09\textheight]{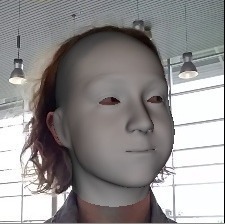}
\end{subfigure}
\begin{subfigure}[t]{0.118\linewidth}
\includegraphics[width = 1.0\textwidth, height = 0.09\textheight]{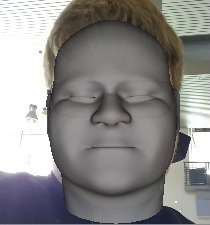}
\end{subfigure}
\begin{subfigure}[t]{0.118\linewidth}
\includegraphics[width = 1.0\textwidth, height = 0.09\textheight]{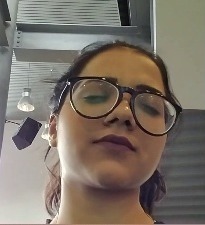}
\end{subfigure}
\begin{subfigure}[t]{0.118\linewidth}
\includegraphics[width = 1.0\textwidth, height = 0.09\textheight]{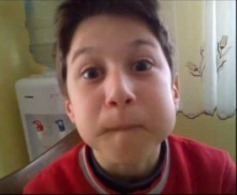}
\end{subfigure}
\begin{subfigure}[t]{0.118\linewidth}
\includegraphics[width = 1.0\textwidth, height = 0.09\textheight]{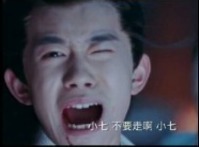}
\end{subfigure}
\begin{subfigure}[t]{0.118\linewidth}
\includegraphics[width = 1.0\textwidth, height = 0.09\textheight]{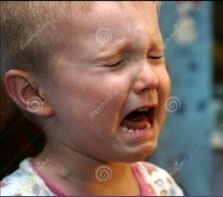}
\end{subfigure}
\begin{subfigure}[t]{0.118\linewidth}
\includegraphics[width = 1.0\textwidth, height = 0.09\textheight]{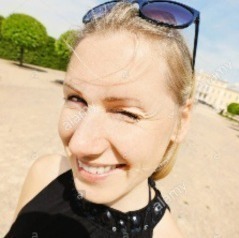}
\end{subfigure}
\begin{subfigure}[t]{0.118\linewidth}
\includegraphics[width = 1.0\textwidth, height = 0.09\textheight]{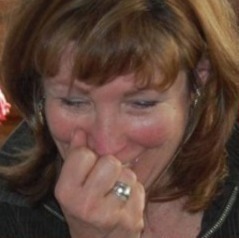}
\end{subfigure}
\begin{subfigure}[t]{0.118\linewidth}
\includegraphics[width = 1.0\textwidth, height = 0.09\textheight]{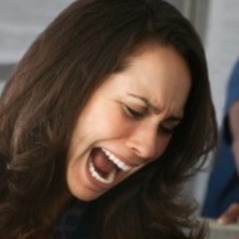}
\end{subfigure}
\begin{subfigure}[t]{0.118\linewidth}
\includegraphics[width = 1.0\textwidth, height = 0.09\textheight]{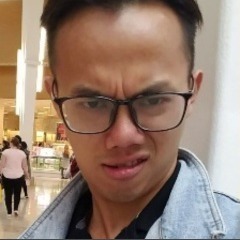}
\end{subfigure}

\begin{subfigure}[t]{0.118\linewidth}
\includegraphics[width = 1.0\textwidth, height = 0.09\textheight]{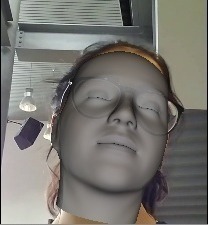}
\end{subfigure}
\begin{subfigure}[t]{0.118\linewidth}
\includegraphics[width = 1.0\textwidth, height = 0.09\textheight]{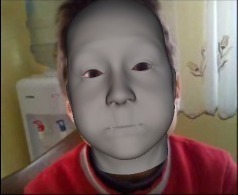}
\end{subfigure}
\begin{subfigure}[t]{0.118\linewidth}
\includegraphics[width = 1.0\textwidth, height = 0.09\textheight]{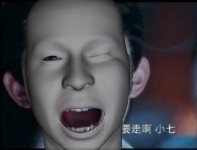}
\end{subfigure}
\begin{subfigure}[t]{0.118\linewidth}
\includegraphics[width = 1.0\textwidth, height = 0.09\textheight]{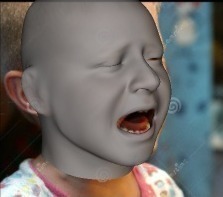}
\end{subfigure}
\begin{subfigure}[t]{0.118\linewidth}
\includegraphics[width = 1.0\textwidth, height = 0.09\textheight]{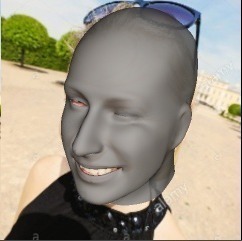}
\end{subfigure}
\begin{subfigure}[t]{0.118\linewidth}
\includegraphics[width = 1.0\textwidth, height = 0.09\textheight]{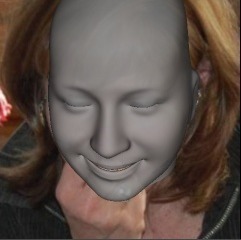}
\end{subfigure}
\begin{subfigure}[t]{0.118\linewidth}
\includegraphics[width = 1.0\textwidth, height = 0.09\textheight]{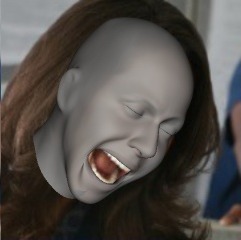}
\end{subfigure}
\begin{subfigure}[t]{0.118\linewidth}
\includegraphics[width = 1.0\textwidth, height = 0.09\textheight]{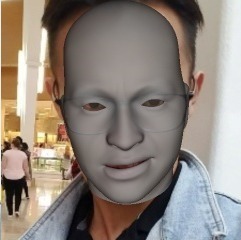}
\end{subfigure}
\caption{More results from our own expression test set using our single face retargeting network.}
\label{fig:result_sfn}
\end{figure*}

\begin{figure*}[h!]
\centering
\begin{subfigure}[t]{0.16\linewidth}
\includegraphics[width = 1.0\textwidth, height = 0.1\textheight]{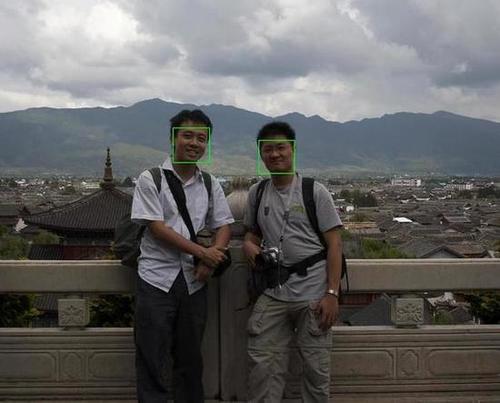}
\end{subfigure}
\begin{subfigure}[t]{0.16\linewidth}
\includegraphics[width = 1.0\textwidth, height = 0.1\textheight]{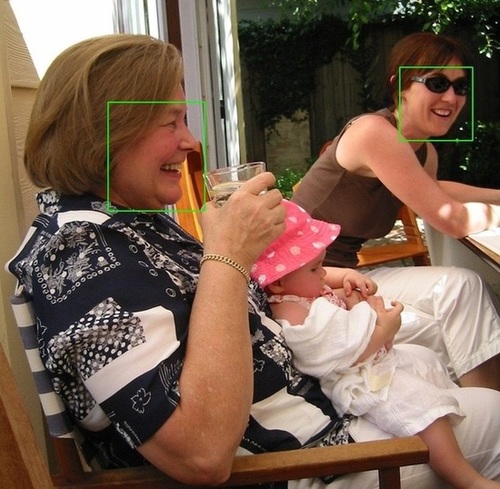}
\end{subfigure}
\begin{subfigure}[t]{0.16\linewidth}
\includegraphics[width = 1.0\textwidth, height = 0.1\textheight]{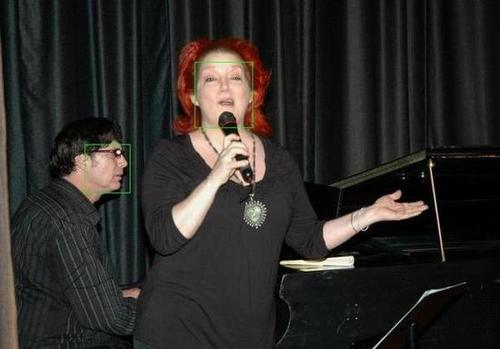}
\end{subfigure}
\begin{subfigure}[t]{0.16\linewidth}
\includegraphics[width = 1.0\textwidth, height = 0.1\textheight]{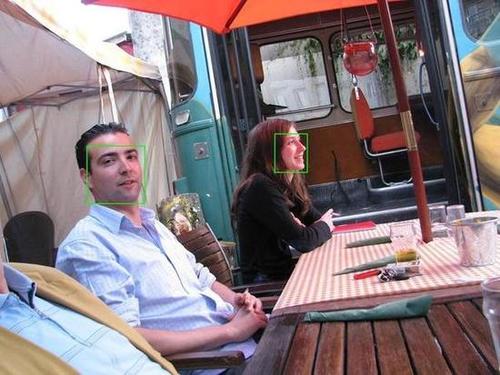}
\end{subfigure}
\begin{subfigure}[t]{0.16\linewidth}
\includegraphics[width = 1.0\textwidth, height = 0.1\textheight]{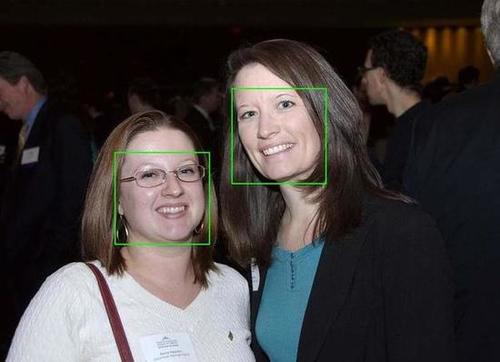}
\end{subfigure}
\begin{subfigure}[t]{0.16\linewidth}
\includegraphics[width = 1.0\textwidth, height = 0.1\textheight]{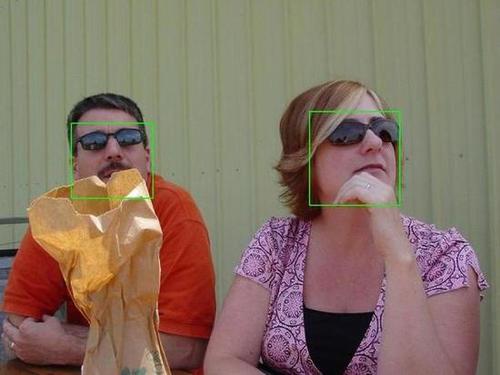}
\end{subfigure}
\begin{subfigure}[t]{0.16\linewidth}
\includegraphics[width = 1.0\textwidth, height = 0.1\textheight]{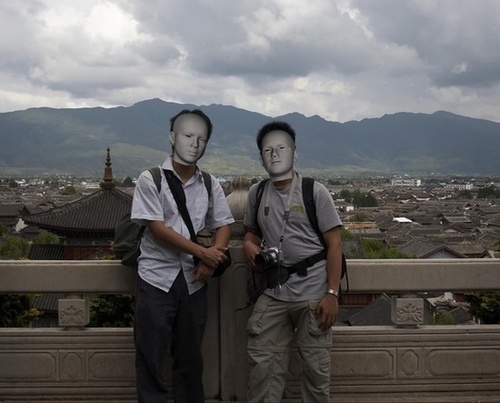}
\end{subfigure}
\begin{subfigure}[t]{0.16\linewidth}
\includegraphics[width = 1.0\textwidth, height = 0.1\textheight]{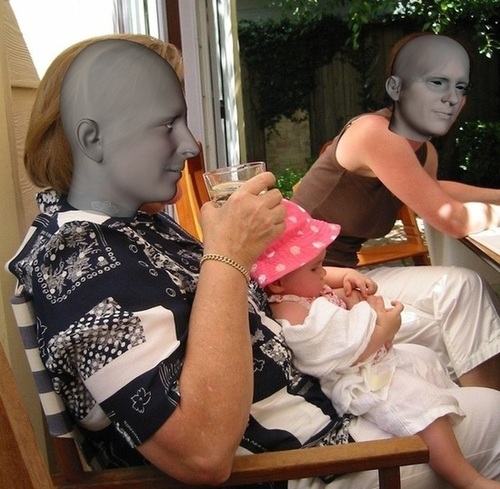}
\end{subfigure}
\begin{subfigure}[t]{0.16\linewidth}
\includegraphics[width = 1.0\textwidth, height = 0.1\textheight]{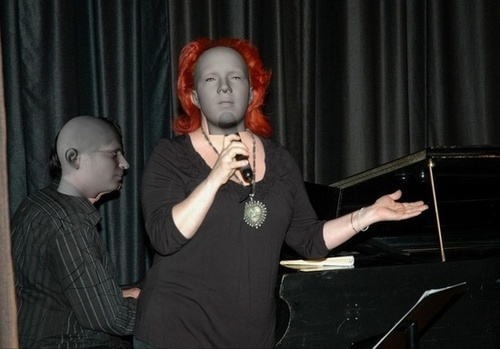}
\end{subfigure}
\begin{subfigure}[t]{0.16\linewidth}
\includegraphics[width = 1.0\textwidth, height = 0.1\textheight]{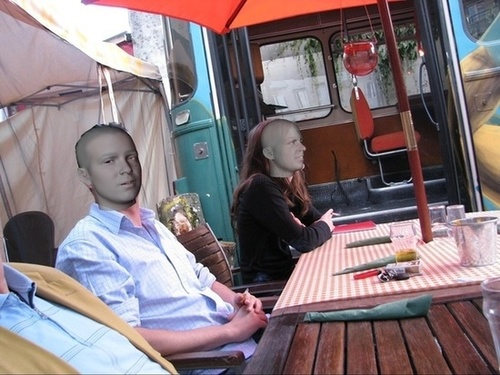}
\end{subfigure}
\begin{subfigure}[t]{0.16\linewidth}
\includegraphics[width = 1.0\textwidth, height = 0.1\textheight]{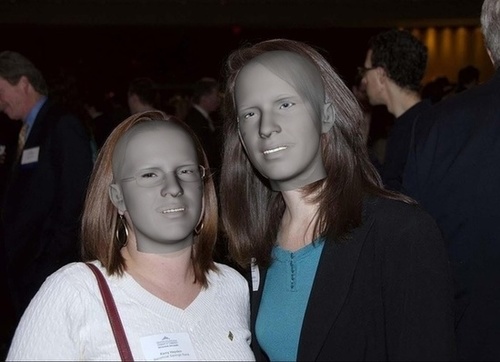}
\end{subfigure}
\begin{subfigure}[t]{0.16\linewidth}
\includegraphics[width = 1.0\textwidth, height = 0.1\textheight]{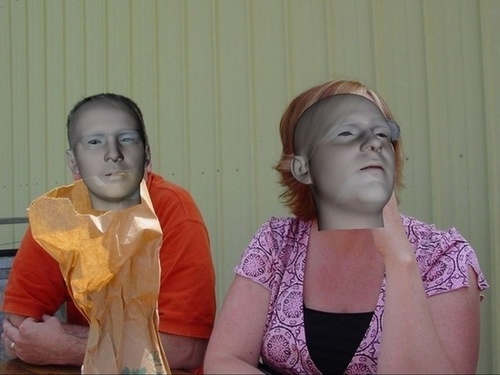}
\end{subfigure}
\begin{subfigure}[t]{0.16\linewidth}
\includegraphics[width = 1.0\textwidth, height = 0.1\textheight]{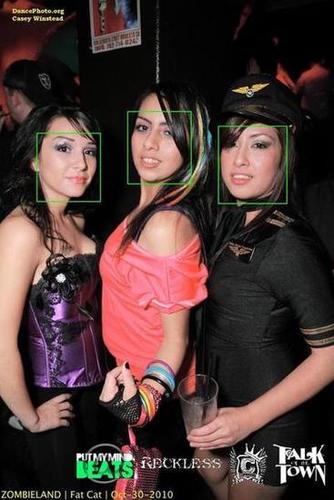}
\end{subfigure}
\begin{subfigure}[t]{0.16\linewidth}
\includegraphics[width = 1.0\textwidth, height = 0.1\textheight]{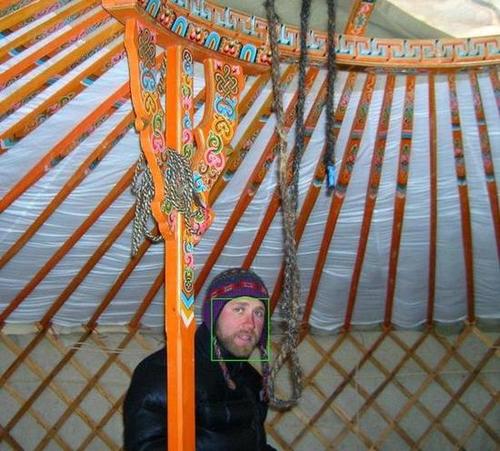}
\end{subfigure}
\begin{subfigure}[t]{0.16\linewidth}
\includegraphics[width = 1.0\textwidth, height = 0.1\textheight]{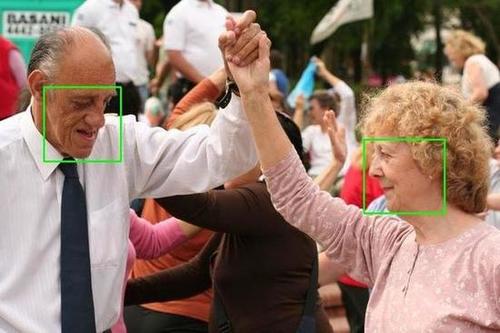}
\end{subfigure}
\begin{subfigure}[t]{0.16\linewidth}
\includegraphics[width = 1.0\textwidth, height = 0.1\textheight]{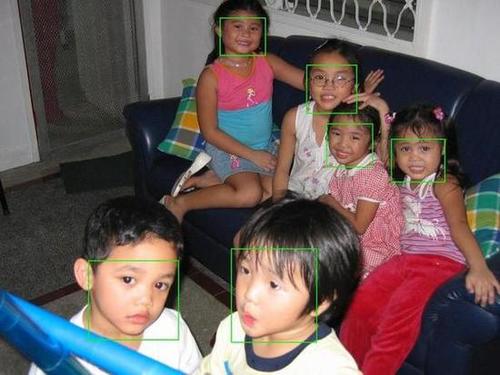}
\end{subfigure}
\begin{subfigure}[t]{0.16\linewidth}
\includegraphics[width = 1.0\textwidth, height = 0.1\textheight]{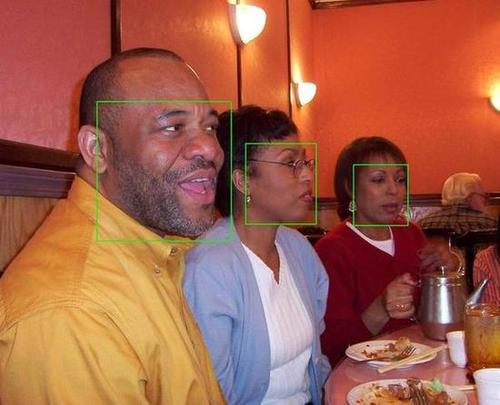}
\end{subfigure}
\begin{subfigure}[t]{0.16\linewidth}
\includegraphics[width = 1.0\textwidth, height = 0.1\textheight]{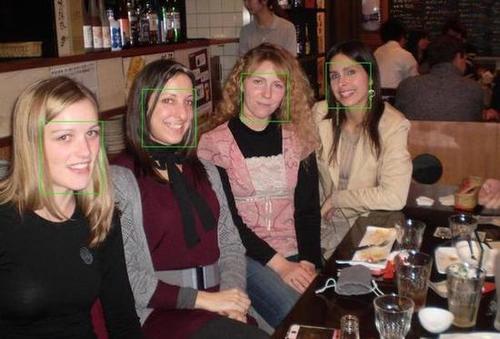}
\end{subfigure}
\begin{subfigure}[t]{0.16\linewidth}
\includegraphics[width = 1.0\textwidth, height = 0.1\textheight]{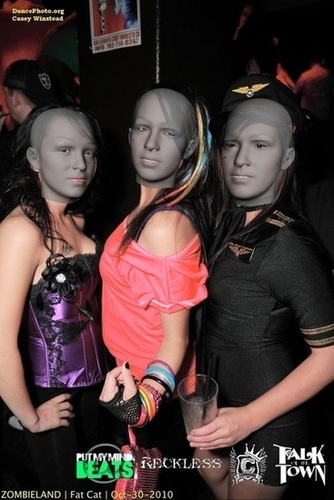}
\end{subfigure}
\begin{subfigure}[t]{0.16\linewidth}
\includegraphics[width = 1.0\textwidth, height = 0.1\textheight]{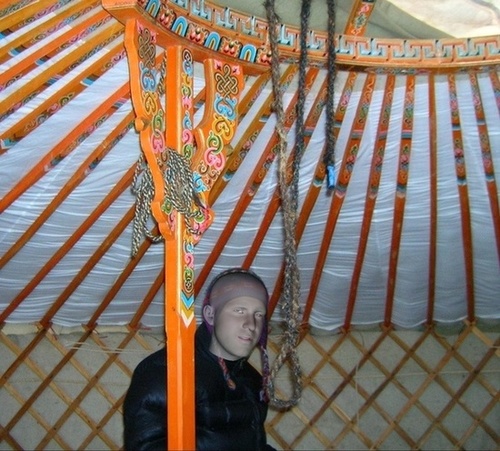}
\end{subfigure}
\begin{subfigure}[t]{0.16\linewidth}
\includegraphics[width = 1.0\textwidth, height = 0.1\textheight]{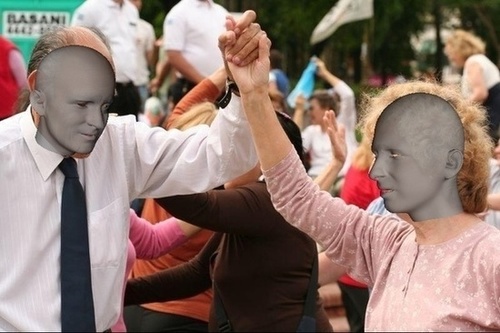}
\end{subfigure}
\begin{subfigure}[t]{0.16\linewidth}
\includegraphics[width = 1.0\textwidth, height = 0.1\textheight]{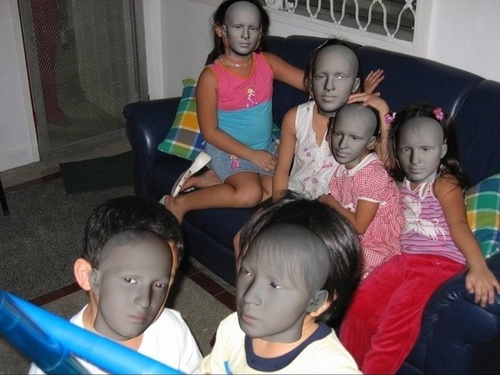}
\end{subfigure}
\begin{subfigure}[t]{0.16\linewidth}
\includegraphics[width = 1.0\textwidth, height = 0.1\textheight]{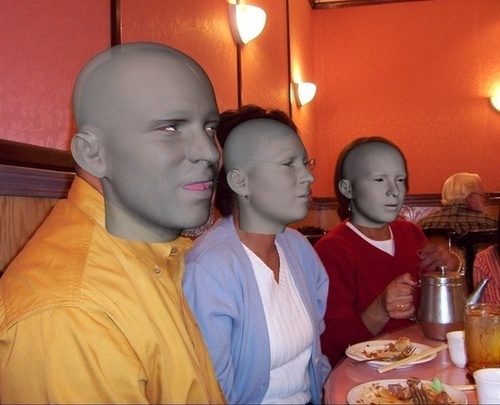}
\end{subfigure}
\begin{subfigure}[t]{0.16\linewidth}
\includegraphics[width = 1.0\textwidth, height = 0.1\textheight]{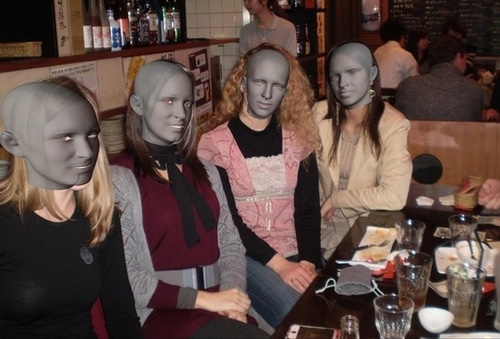}
\end{subfigure}
\caption{More results from AFW dataset \cite{afwdatapaper} using our joint detection and retargeting model.}
\label{fig:result_mfn}
\end{figure*}

\subsection{Accuracy of Face Detection}
Our network can detect multiple small faces of reasonable size even though it is not trained on images more than 20 faces. Fig. \ref{fig:facedetection} shows our network outputs for an image with more than 20 faces in the AFW dataset.

\subsection{Video Results}
The performance of our networks on videos is shown in a video\footnote{\url{https://homes.cs.washington.edu/~bindita/cvpr2019.mp4}}. In the first half, we show the results of retargeting from a single face video to a generic 3D human face model using our single face retargeting network. The face bounding box for the current frame is obtained from the boundaries of the 2D landmarks predicted in the previous frame. In the second half, we show the results of retargeting with videos having multiple faces (Music video dataset \cite{musicvideo}) using our multi-face retargeting network (only frames with multiple faces are shown).

\end{document}